*Technical Note*

# Pano-GAN: A Deep Generative Model for Panoramic Dental Radiographs

Søren Pedersen [1,†], Sanyam Jain [2,*,†], Mikkel Chavez [1,†], Viktor Ladehoff [1,†], Bruna Neves de Freitas [2,3] and Ruben Pauwels [2,*]

1 Bachelor's Degree Programme in Data Science, Aarhus University, Nordre Ringgade 1, 8000 Aarhus, Denmark; 202308323@post.au.dk (S.P.); 202109869@post.au.dk (M.C.); 202108007@post.au.dk (V.L.)
2 Department of Dentistry and Oral Health, Aarhus University, Vennelyst Boulevard 9, 8000 Aarhus, Denmark; brunanf@aias.au.dk
3 Aarhus Institute of Advanced Studies, Aarhus University, Høegh-Guldbergs Gade 6B, 8000 Aarhus, Denmark
* Correspondence: sanyam.jain@dent.au.dk (S.J.); ruben.pauwels@dent.au.dk (R.P.)
† These authors contributed equally to this work.

**Abstract:** This paper presents the development of a generative adversarial network (GAN) for the generation of synthetic dental panoramic radiographs. While this is an exploratory study, the ultimate aim is to address the scarcity of data in dental research and education. A deep convolutional GAN (DCGAN) with the Wasserstein loss and a gradient penalty (WGAN-GP) was trained on a dataset of 2322 radiographs of varying quality. The focus of this study was on the dentoalveolar part of the radiographs; other structures were cropped out. Significant data cleaning and preprocessing were conducted to standardize the input formats while maintaining anatomical variability. Four candidate models were identified by varying the critic iterations, number of features and the use of denoising prior to training. To assess the quality of the generated images, a clinical expert evaluated a set of generated synthetic radiographs using a ranking system based on visibility and realism, with scores ranging from 1 (very poor) to 5 (excellent). It was found that most generated radiographs showed moderate depictions of dentoalveolar anatomical structures, although they were considerably impaired by artifacts. The mean evaluation scores showed a trade-off between the model trained on non-denoised data, which showed the highest subjective quality for finer structures, such as the *mandibular canal* and *trabecular bone*, and one of the models trained on denoised data, which offered better overall image quality, especially in terms of *clarity and sharpness* and *overall realism*. These outcomes serve as a foundation for further research into GAN architectures for dental imaging applications.

## 1. Introduction

Artificial intelligence (AI) is revolutionizing the field of dentistry, offering novel tools that can automate diagnostic processes, streamline clinical workflows, and enhance patient outcomes [1]. Over the past decade, AI-driven applications have shown significant progress, particularly within image analysis, where tasks such as segmentation, classification, and anomaly detection have traditionally required manual input from dental professionals [2,3]. These AI technologies, involving machine learning (ML) and especially its subset, deep learning (DL), are now capable of identifying intricate patterns within dental radiographs, providing faster and more accurate assessments of dental conditions. As a result, AI applications have the potential to significantly reduce human errors, improve the efficiency of dental diagnostics, and contribute to personalized, preventive, and predictive dentistry [4]. In addition, AI can be integrated into 2D or 3D image processing workflows for the segmentation of hard tissue and other structures, uni- or multi-modal registration, image enhancement, and tomographic reconstruction [5,6]. Despite these advancements, however, the adoption of AI in routine clinical practice has been relatively slow, hindered by challenges such as limited training data, ethical concerns, and the lack of transparency in AI models [3].

A crucial barrier to the implementation of AI in dental radiography is the scarcity of high-quality annotated data, which are essential in training robust AI models. Dental radiographs, such as panoramic radiographs, contain complex anatomical structures that require precise annotation, and collecting and curating large-scale datasets is both time-consuming and resource-intensive [7]. Moreover, the ethical implications of using AI-generated data in healthcare are significant and must be carefully considered [3]. Issues related to data privacy, patient consent, and data sharing between institutions complicate the availability of large datasets for the training of AI models in dentistry [2]. In recent years, generative adversarial networks (GANs) [8] have emerged as a powerful solution to overcome these limitations by supplementing real-world data with synthetic samples. GANs consist of two competing neural networks: a generator, which creates synthetic images, and a discriminator, which evaluates their authenticity. Through an iterative process, the generator improves its ability to produce realistic images that are increasingly indistinguishable from real data. GANs, by generating diverse synthetic yet realistic data, could supplement existing datasets and address the data scarcity and patient privacy problems in the medical and dental fields. The ability to generate realistic synthetic data can also be useful in clinical education, as these data can be tailored to reflect a wide range of anatomical variations and pathologies. This customization would enable a more individualized teaching approach, allowing us to target certain weaknesses or focus areas in an automated and student-specific manner.

In this study, we utilize a deep convolutional GAN (DCGAN) architecture, enhanced by the Wasserstein loss with a gradient penalty (WGAN-GP), to generate synthetic panoramic radiographs (Figure 1). The Wasserstein GAN (WGAN) introduces stability to the GAN training process by addressing common issues such as mode collapse and training instability, making it particularly well suited for the generation of high-quality medical images [9]. Panoramic radiographs are selected as a use case, as they are commonly used in dentistry for the diagnosis of dental and maxillofacial conditions. Furthermore, they are relatively standardized in terms of anatomical depiction, showing the entire dentoalveolar region centralized on the image, along with adjacent structures such as the maxillary sinus and temporomandibular joint. Previous work involving DL and panoramic radiographs has focused on automatic segmentation and detection models, which have been used to interpret panoramic X-rays for various clinical purposes [10–13]. Most of these studies involve datasets that are limited in size, showing the potential of making these models more comprehensive and generalizable through the addition of synthetic data. The use of synthetic data in DL model training is where the main contribution of the current study can be found, although its exploratory nature precludes an evaluation of the effect on DL model training at this point.

In the following sections, we present the related work, methods, results, and discussion. We finally conclude the paper with the results, including generated images and evaluation plots.

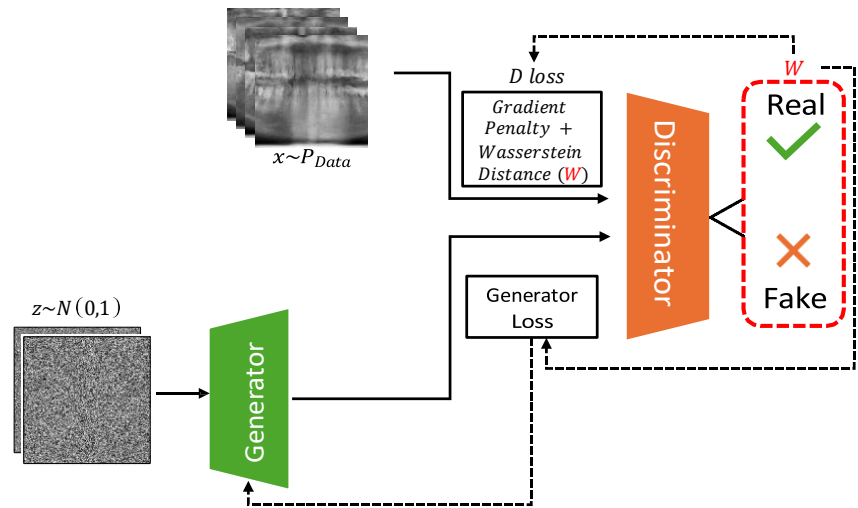

**Figure 1.** An overview of WGAN-GP. The generator receives input noise and produces samples, which are compared against real data from the dataset in the discriminator. The figure illustrates the calculation of the critic loss (D loss) by real, fake, and gradient penalty terms, along with the generator loss (G loss). The gradient penalty term is highlighted alongside the Wasserstein distance in a separate box.

## 2. Related Work

GANs have been extensively studied for image-to-image tasks across various fields, including medical imaging and dentistry. The primary applications of GANs in medical imaging are found in image enhancement, including denoising [14,15], super-resolution [16], and the inpainting of sparse raw data for tomographic imaging [17]. Furthermore, inter-modality conversion using GANs has been explored, including conversion between computed tomography, cone-beam computed tomography, and magnetic resonance imaging [18,19]. GANs have also been suggested as transformation regularizers in image registration [20]. However, when used to generate entire synthetic images in particular, GANs suffer from instability during training, frequently encountering issues such as mode collapse and vanishing gradients. To address these shortcomings, WGAN was proposed [9], which introduced a more robust training objective by using the Wasserstein distance (also known as the earth mover's distance) as a measure of how different the generated data distribution is from the real data distribution.

### 2.1. Wasserstein GAN with Gradient Penalty (WGAN-GP)

One of the key improvements to WGAN was introduced by Gulrajani et al. [21]. While the original WGAN used weight clipping to enforce the Lipschitz continuity condition (required for the Wasserstein distance), this method introduced new problems, such as poor optimization performance. WGAN-GP solves this by applying a gradient penalty to the discriminator's loss function, ensuring that the gradients of the discriminator are close to 1 around the real and generated data points, which is necessary for good training stability. The WGAN-GP loss function is defined as

$$\mathsf{L}_D = \mathbb{E}_{x^\sim \sim \mathrm{P}_g}[D(x^\sim)] - \mathbb{E}_{x \sim \mathrm{P}_r}[D(x)] + \lambda \mathbb{E}_{x^\wedge \sim \mathrm{P}_{x^\wedge}}\left[(\|\nabla_{\hat{x}} D(\hat{x})\|_2 - 1)^2\right] \quad (1)$$

where $x$ is sampled from the real data distribution $\mathrm{P}_r$, $x^\sim$ is sampled from the generated data distribution $\mathrm{P}_g$, and $x^\wedge$ is sampled from a uniform distribution along the straight lines between the points in $\mathrm{P}_r$ and $\mathrm{P}_g$. The third term, $\lambda \mathbb{E}_{x^\wedge}\left[(\|\nabla_{x^\wedge} D(x^\wedge)\|_2 - 1)^2\right]$, enforces the gradient penalty, with $\lambda$ controlling the strength of the penalty.

This gradient penalty replaces weight clipping and ensures that the discriminator remains within the 1-Lipschitz constraint, which is crucial in measuring the Wasserstein distance effectively. By using the gradient penalty, WGAN-GP improves the training stability and convergence, allowing it to generate higher-quality images more consistently than other GANs.

### 2.2. Relevance of WGAN-GP to Our Work and Comparison with Other GANs

In this study, we have selected WGAN-GP due to its robustness in training and ability to yield high-dimensional, complex image data; this is a necessity for the generation of dental radiographs containing intricate structures. The stability of WGAN-GP allows the generator to learn to produce such detailed features without suffering from common issues associated with traditional GAN architectures, including mode collapse and poor gradient flows. Specifically, the incorporation of a gradient penalty helps to mitigate the problems of vanishing or exploding gradients, which are detrimental to the fine details needed in medical imaging. In contrast, traditional GANs and even WGANs with weight clipping often struggle to produce stable gradients, especially in datasets with high variability, such as dental radiographs. Various other GAN variants, like least-squares GAN [22], reduce the vanishing gradient problem but fail to effectively address mode collapse. Similarly, DCGANs [23] introduce convolutional layers for enhanced image generation yet still face instability in complex tasks. Other architectures, such as progressive GANs [24], may enhance the image quality further but often require more computational resources and longer training times.

## 3. Methods

**Data:** The dataset consisted of 2322 dental panoramic radiographs in PNG format from the MICCAI-DENTEX (Dental Enumeration and Diagnosis on Panoramic X-Rays) Challenge [25]. The dataset is made available under a CC BY-SA 4.0 License. It consists of images from various panoramic radiography machines (from different manufacturers), varying in image quality and resolution. A custom transformation pipeline was implemented to preprocess the X-ray images before they were fed into the model. This involved a custom cropping function focusing on the bottom-center part of each image, representing the dentoalveolar region. In other words, the aim of this exploratory study was to generate partial panoramic radiographs containing the teeth and adjacent bone structures, but not the temporomandibular joint, maxillary sinus, or other anatomical regions found in a full-sized panoramic radiograph. After cropping, the cropped images were resized and converted to 8-bit grayscale with normalized pixel values, which helped in stabilizing the training process.

**Architecture:** WGAN-GP was implemented to address the challenges of training stability and image quality in our project. The core of the WGAN-GP architecture is based on the Wasserstein distance $W(P_r, P_g)$, which is defined as

$$W(P_r, P_g) = \inf_{\gamma \in \Pi(P_r, P_g)} \mathbb{E}_{(x,y) \sim \gamma}[\|x - y\|], \tag{2}$$

where $\Pi(P_r, P_g)$ denotes the set of all joint distributions with specified marginals $P_r$ (real data distribution) and $P_g$ (generated data distribution). A key aspect of our implementation is the gradient penalty, which is computed as

$$L_{GP} = \mathbb{E}_{\hat{x} \sim P_{\hat{x}}} \left[ (\|\nabla_{\hat{x}} D(\hat{x})\|_2 - 1)^2 \right], \tag{3}$$

where $D$ is the critic network, and $\hat{x}$ is the interpolated image between the real and generated samples. This penalty encourages the gradients of the critic to have a norm close to 1, thereby stabilizing the training.

The architectures of both the generator and discriminator are crucial in achieving high-quality image generation. The discriminator is implemented as a convolutional neural network consisting of several layers, where each layer applies convolutional operations followed by instance normalization (IN) and leaky ReLU activation. Specifically, the network progressively reduces the spatial dimensions of the input image $I \in \mathbb{R}^{C \times H \times W}$ to produce a single scalar output representing the critic's score $D(I)$:

$$D(I) = \text{Disc}(\text{Conv2D}(\text{LeakyReLU}(I))), \qquad (4)$$

where Disc denotes the sequential layers of the discriminator. The use of IN is essential to ensure a focus on local patterns and textures for individual samples, rather than style-related statistics. It is particularly useful for heterogeneous datasets; in our study, while the images were homogeneous in terms of modality, there was considerable variation in style, i.e., between radiographs from different panoramic radiography units. IN also helps to stabilize training through a more consistent gradient flow. As for leaky ReLu, it was found that its general ability to prevent permanently inactive neurons, improve and balance the gradient flow, and allow detailed feature learning was highly beneficial for a WGAN set-up.

The generator, on the other hand, is designed to upsample a noise vector $z \in \mathbb{R}^{C_{noise} \times 1 \times 1}$ into a high-resolution image $I' \in \mathbb{R}^{C_{img} \times 256 \times 256}$ using transposed convolutional layers. For all hidden layers, batch normalization (BN) and ReLU activation were applied to enhance the training stability and efficiency. BN helps to stabilize the gradient flow, ensures consistent feature scaling, facilitates meaningful exploration of the latent space, and accelerates convergence during training. ReLU, a widely used activation function in deep learning, is particularly useful within the context of a generator by enabling non-linear feature learning, ensuring computational efficiency, mitigating vanishing gradient issues, and creating sparsity within the network. The final output is generated using the following operation:

$$I' = G(z) = \text{ConvT2D}(\text{ReLU}(\text{BN}(z))), \qquad (5)$$

culminating in a tanh activation function to produce images in the range of $[-1, 1]$. The implemented generator and discriminator networks are shown in Figure 2.

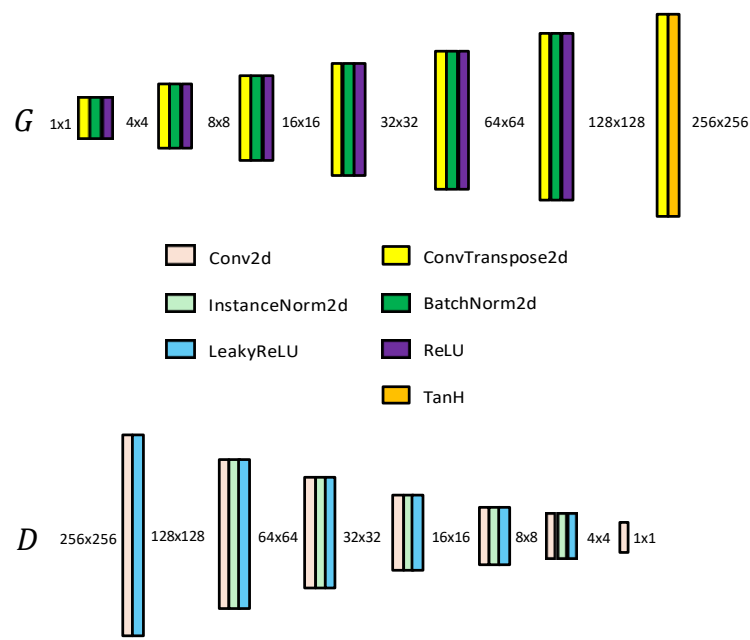

**Figure 2.** Generator $G(z)$ and discriminator $D(I)$ networks used in methodology.

**Implementation:** We implemented four different models using the same overall architecture shown above. One essential difference between the models was in the use of denoising prior to training. Model 1 was trained on the original dataset without any denoising applied, while Models 2, 3, and 4 were trained using images denoised with anisotropic diffusion (AD) [26]. The denoising level applied to Models 2–4 followed the default implementation of AD for FIJI [27]. The training set for Models 2–4 was also narrowed down to 1800 images by manually screening the initial dataset for images of poor quality. Furthermore, the number of critic iterations was varied in an attempt to optimize the balance between generator and discriminator performance. Finally, the number of epochs was varied. The configuration for the models is outlined in Table 1.

**Table 1.** Experimental configuration for Models 1 through 4, including input dimensions (W = width, H = height), critic iterations, number of training epochs, and use of denoising.

| M | (W, H) | Critic | Epochs | Denoise |
|---|---|---|---|---|
| M1 | $256 \times 256$ | 2 | 550 | No |
| M2 | $256 \times 256$ | 1 | 150 | Yes |
| M3 | $256 \times 256$ | 4 | 250 | Yes |
| M4 | $256 \times 256$ | 5 | 100 | Yes |

**Objective evaluation:** We implemented two metrics for the evaluation of the generated images and compared them with the training samples. (1) The Fréchet inception distance (FID) [28] evaluates the quality of generated images by measuring the distance between the feature distributions of real and generated images, as extracted by a pretrained inception network. The FID values are lower when the generated images closely match the real image distribution, indicating better quality and diversity. This metric captures both feature-level similarity and distributional overlap, making it suitable for assessing generative models. (2) t-Distributed Stochastic Neighbor Embedding (t-SNE) [29] is a dimensionality reduction technique used to visualize high-dimensional data in 2D or 3D space while preserving the local structure. In this case, t-SNE was applied to the ResNet50-extracted [30] features of real and generated images to visualize their separability and clustering in a lower-dimensional space. Overlapping clusters indicate similarity between the real and generated features, while distinct bulging clusters suggest clear separability and feature differences.

**Expert evaluation:** For each of the four models, 25 generated radiographs were evaluated by a dentist with 9 years of experience, including a Master's degree in Oral Rehabilitation and a specialty course in Dentomaxillofacial Radiology. The evaluation was performed on an interleaved pooled dataset of 100 generated images, in batches of 20 images per session to avoid fatigue. A specific scoring system was developed for this study, using a five-point scale for twelve criteria, ranging from overall realism to specific anatomical features. The criteria were

- overall realism;
- clarity and sharpness;
- the tooth anatomy—depiction of crowns and roots;
- the jaw and bone structure—depiction of the cortical bone and trabecular bone;
- the alignment and symmetry of the teeth and jaws, focusing on the bone shape and the visualization of the occlusal space;
- the absence of artifacts that detract from the realism; and
- other landmarks—visualization of the mandibular canal and hard palate.

Each criterion was assessed in terms of its representation with regard to real-world panoramic radiographs. The scoring system was defined from 1 to 5, where 1 indicated a poor or highly unrealistic outcome and 5 signified excellent or highly realistic qualities.

Prior to the observation, the expert was familiarized with the scoring system by applying it to a separate set of 10 generated images.

## 4. Results

### *4.1. Objective Metrics of Synthetic Radiographs*

The FID scores for 25 generated images (F) per model vs. 2000 real images (R) are shown in Table 2. To put the FID scores into perspective, a comparison of 25 R vs. 2000 R was included, along with 25 Gaussian noise images (G) vs. 2000 R. The G images were generated using $256 \times 256$ pixels, $\mu = 128$ and $\sigma = 64$. The FID values for F vs. R were considerably higher than for R vs. R, but lower than for G vs. R. This indicates a gap in similarity between the generated and real images, while also showing that the generated images exhibited the structural patterns of actual panoramic radiographs. Figure 3 shows example images for R, F, and G, as well as a t-SNE plot. In line with the FID findings, the t-SNE visualization reveals distinct clustering patterns for the R, F, and G groups. The relative proximity between the R and F clusters indicates partial feature alignment, confirming that the generated images trend towards realism, while remaining distinct and thus falling short of true real-life detail.

**Table 2.** FID scores compared with real (R), Gaussian noise (G), and fake (F) images from each respective model (M).

| Data | Reference | FID Score |
| --- | --- | --- |
| 25R | 2000 R | 138.4 |
| 25F-M1 | 2000 R | 364.5 |
| 25F-M2 | 2000 R | 312.8 |
| 25F-M3 | 2000 R | 342.4 |
| 25F-M4 | 2000 R | 320.3 |
| 25G | 2000 R | 658.2 |

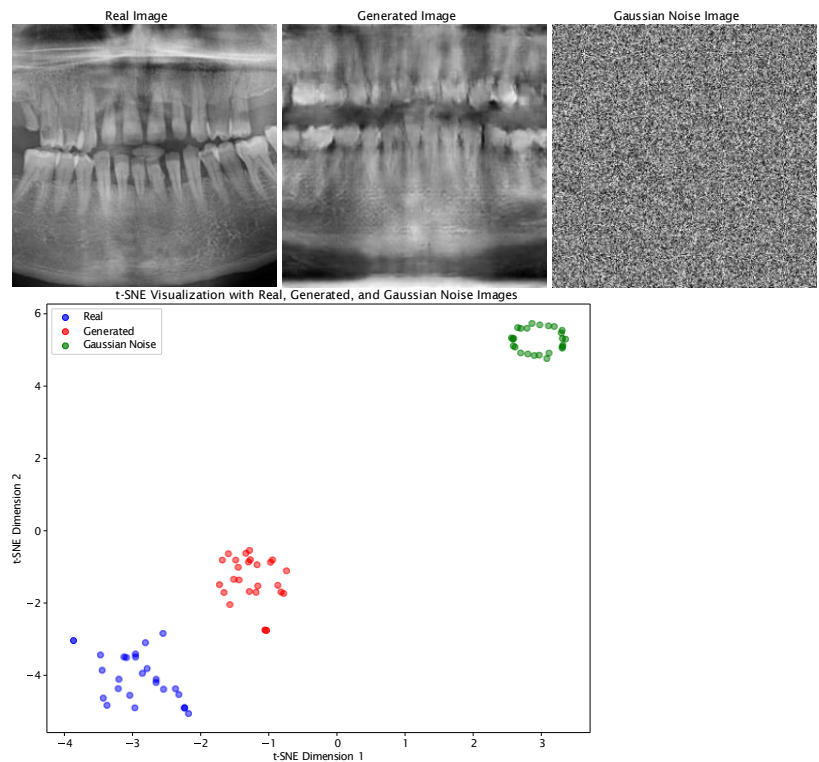

**Figure 3.** Examples of real (R), fake (F), and Gaussian noise (G) images. The t-SNE plot compares the feature embedding for R (blue cluster), F (red cluster), and G (green cluster) images.

### 4.2. Expert Evaluation of Synthetic Radiographs

The overall distribution of the expert scores is shown in Table 3. Model 1 showed the highest average score for four criteria, whereas Model 2 showed the highest score for eight criteria (including one tied score with Model 4). A statistical evaluation of the scores is presented in the Appendix A.

**Table 3.** Expert evaluation scores, ranging from 1 to 5. The model with the highest average score for each criterion is shown in bold. Models 1 and/or 2 show the best average performance for most dental and image quality attributes. Abbreviations: M = Model, OR = Overall Realism, CS = Clarity and Sharpness, CB = Contrast and Brightness, CR = Crowns, RT = Roots, CTB = Cortical Bone, TB = Trabecular Bone, BS = Bone Shape, OS = Occlusal Space, AA = Absence of Artifacts, MC = Mandibular Canal, HP = Hard Palate.

| M | OR | CS | CB | CR | RT | CTB | TB | BS | OS | AA | MC | HP |
|---|---|---|---|---|---|---|---|---|---|---|---|---|
| M1 | 2.44 | 1.68 | 2.84 | **2.48** | 2.00 | 2.48 | **3.36** | 2.48 | 2.76 | 1.48 | **1.64** | **3.20** |
| M2 | **2.72** | **1.96** | **2.88** | 2.32 | **2.20** | **2.76** | 2.80 | **3.16** | **3.04** | **1.64** | 1.48 | 3.04 |
| M3 | 1.88 | 1.52 | 2.40 | 2.04 | 1.96 | 1.72 | 1.80 | 2.56 | 2.24 | 1.12 | 1.60 | 3.04 |
| M4 | 2.32 | 1.76 | 2.60 | 2.40 | **2.20** | 2.00 | 1.96 | 2.56 | 2.64 | 1.44 | 1.20 | 3.04 |

Based on these findings, the scores for Model 1 and 2 are presented in more detail. As stated earlier, the main difference between these models is whether they were trained on original radiographs (Model 1) or denoised ones (Model 2). Boxplots showing the distribution of the expert scores are shown in Figure 4. It can be seen that, for most metrics, the scores ranged between 1 and 5; the most notable exceptions were for *clarity and sharpness*, *roots*, and the *absence of artifacts* (for which the scores did not surpass 3). Furthermore, differences in the width of the score distributions could be observed for various criteria, showing more consistent outputs for one of the models (either better or worse). Figure 5 shows a radar plot of the average scores for the two models, allowing for a convenient comparison of their relative performance. It can be seen that Model 1 mainly outperformed Model 2 for *trabecular bone*; this is likely related to the denoising used for the input data of Model 2, causing some degree of blurring of small structures such as trabeculae. On the other hand, Model 2 showed higher scores for the *bone shape* in particular and for *cortical bone*, *clarity and sharpness*, *the occlusal space*, and *overall realism* to some extent; this indicates that the absence of noise allowed for the improved generation of larger structures, as well as representing the typical image quality of a panoramic radiograph (somewhat) better.

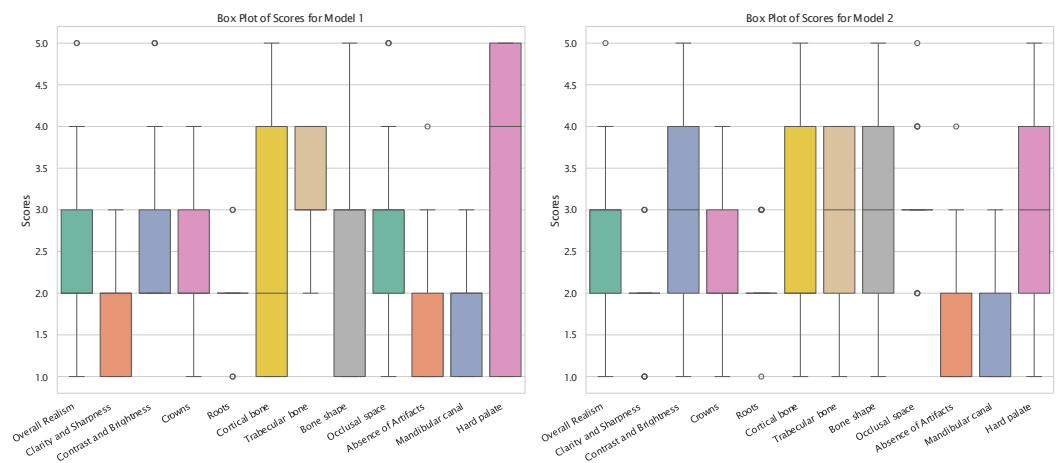

**Figure 4.** Boxplots of observer scores for Models 1 and 2.

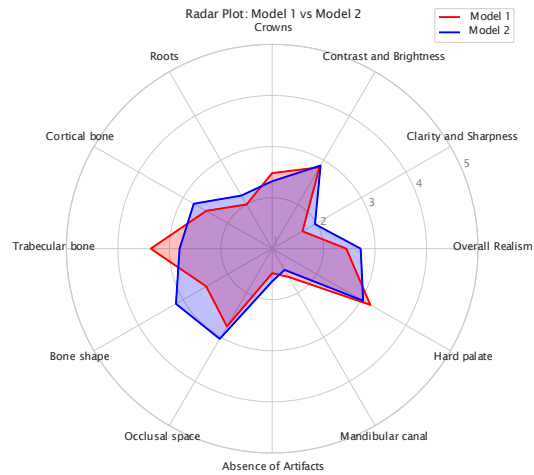

**Figure 5.** Radar plots for Models 1 (red) and 2 (blue).

Figures 6 and 7 show the best and worst generated images, respectively, as determined by the average scores for all twelve criteria. The best images show a reasonable depiction of both the overall morphology and certain details, although not at the level of contemporary real-world panoramic radiographs. The worst images display various issues, including poor overall image quality and manifestations of extra rows of tooth crowns.

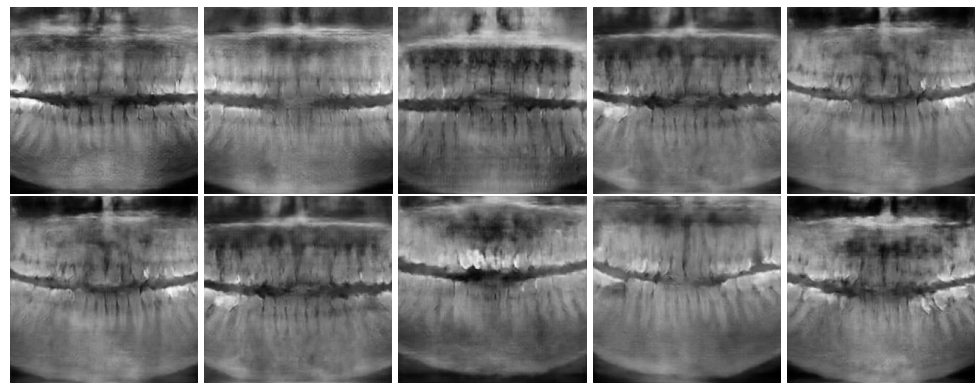

**Figure 6.** *Best images* generated using Models 1 (top row) and 2 (bottom row).

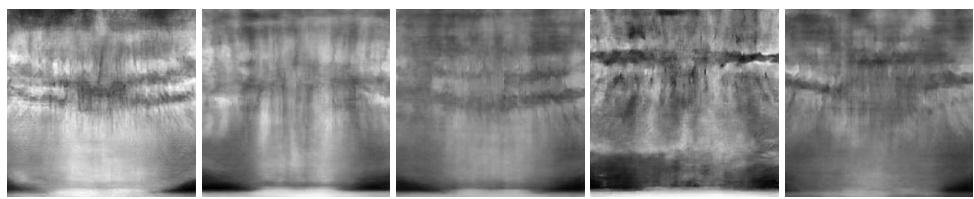

**Figure 7.** *Worst images* among all model variants, showing poor overall anatomical depiction and severe artifacts.

### 4.3. Summary of Other Exploratory Tests

Apart from the models shown here, other exploratory tests are described at https://github.com/ViktorLaden/DataProjectGAN (accessed on 29 January 2025), along with a few illustrative examples. A standard GAN was applied first, but this resulted in highly imbalanced training, leading us to focus on WGAN-GP as an architecture, a choice supported by the theoretical considerations described in the Section 2. Next, we attempted to generate complete panoramic radiographs, containing not only teeth but adjacent anatomical structures such as the temporomandibular joint and maxillary sinus as well. This proved to be challenging, as generating images with an aspect ratio equal to that of a panoramic radiograph caused excessive blurring. Generating high-resolution images with detailed features across a large area like a panoramic radiograph requires significant computational resources and data. Thus, any limitations in the model capacity or training data can lead to increased blurring. On the other hand, resizing the radiographs to a 1:1 aspect ratio without cropping inevitably caused deformation, which would have required a second DL model to convert the generated images into realistic aspect ratios. These were the main factors leading us to focus on squarely cropped images of the dentoalveolar region at this point.

## 5. Discussion

In this exploratory study, the potential of GANs to generate synthetic dental panoramic radiographs was investigated by implementing a DCGAN with the Wasserstein loss and a gradient penalty (WGAN-GP). Rather than generating entire radiographs, the training data were cropped to the dentoalveolar region to ensure that the WGAN did not focus on areas with lower diagnostic value. The dentoalveolar region is the main region of interest in dental radiology; it contains a wide variety of anatomical structures and other features, such as radiolucent and radiopaque pathoses, dental restorations, etc.

The ultimate aim of this research was to produce full-sized synthetic images that could support predictive modeling and educational purposes, in order to address the critical issue of data scarcity in dental research and teaching. Although objective and subjective evaluations indicated that the generated images did not meet the standards of realism and clarity for real-life implementation, the project's findings emphasize the importance of GANs as a viable tool in enhancing the availability of medical imaging data. This study can be considered as an initial endeavor that is likely to lead to further advancements in this field of research.

Using a dedicated scoring system for synthetic panoramic radiographs, the overall evaluation showed that the images achieved varied levels of realism, with many exhibiting a moderate degree of clarity and anatomical detail, but also indicating several areas for improvement. Scores were assigned based on the specific features visible in each image. The results highlight both the strengths and limitations of the model's ability to generate images that closely resemble true panoramic radiographs. At this stage, an observation approach in which a clinical expert acts as a discriminator who judges whether images are real or fake was not yet considered, as this would require a significant improvement in the generated images' quality.

These findings suggest that the further optimization of the GAN architecture and training methodology is necessary to achieve higher fidelity in synthetic image generation. This will likely include a substantial increase in the training sample size, potentially requiring a multicenter approach. Particular focus should be placed on the fact that panoramic radiographs from different manufacturers exhibit varying image quality. While variations in brightness and contrast can be normalized to some extent, differences in sharpness and noise cannot be overlooked. Furthermore, the anatomical coverage for panoramic radiographs from different manufacturers varies to some extent; while all of these images cover, and are centered on, the dentoalveolar region, the lateral and cranial extent of the radiograph varies. This was not relevant in the current study, as the input data were cropped to the dentoalveolar region, but it will be of importance when full-sized panoramic radiographs are generated.

To enable higher-resolution image generation, future research could consider the use of progressive growing strategies [24], hierarchical structures [31], or multi-scale generators [32]. An alternative approach would be the use of super-resolution techniques for refinement after generation [33]. When using diffusion models, advanced noise schedules or variance-preserving methods may result in improved stability and detail for high-resolution outputs [34,35]. Furthermore, pre-trained high-resolution feature extractors for perceptual loss [36,37] or adaptive instance normalization [38] could improve the realism of full-scale generated radiographs at a reasonable computational cost.

Future work should also focus on approaches towards synthetic radiography that are less naive, i.e., using pre-labeled data. For example, a panoramic radiograph can be segmented into the teeth (with each tooth being numbered and restorations being marked), bone(s), mandibular canal, maxillary sinus, etc. This would split the generation into a number of separate tasks, possibly allowing for template-based image-to-image generation or even text-to-image generation. Either approach would also allow for a degree of customization, including the insertion of various types of pathosis (e.g., caries, periapical lesions, root fractures, periodontal bone loss). Once this stage is reached, synthetic radiography can be implemented in personalized dental training, as well as predictive model development. At this point, the ethical considerations surrounding the use of synthetic data in clinical settings become crucial, as these data should not introduce biases or compromise the diagnostic (or other) accuracy. Once synthetic radiography reaches a level at which it can be integrated into the development of AI tools for diagnosis, segmentation, or other tasks, further work will be needed to evaluate the optimal balance between synthetic and real data for training as well as benchmarking.

## 6. Conclusions

The use of a WGAN-GP for the generation of synthetic panoramic radiographs shows promise. We utilized two evaluation metrics, with their respective observations being as follows: (1) low FID scores for real–real, moderate for fake–real, and very high for Gaussian–real images and (2) t-SNE observations indicating that real and fake images are closer in the feature space compared with Gaussian noisy images. Furthermore, based on the expert evaluation scores, various nuances can be seen within the models' performance in terms of anatomical depiction. The most likely bottleneck for model performance within this study was the training sample size; furthermore, the use of training data that are anatomically (and pathologically) labeled should be considered in future work. Finally, the use of complex generative techniques like diffusion models suitable for high-resolution image generation should be considered.

**Author Contributions:** Conceptualization, S.P., S.J., M.C., V.L. and R.P.; methodology, S.P., S.J., M.C., V.L., B.N.d.F. and R.P.; software, S.P., M.C. and V.L.; validation, S.J., B.N.d.F. and R.P.; investigation, S.P., S.J., M.C., V.L., B.N.d.F. and R.P.; resources, R.P.; data curation, S.P., M.C., V.L. and R.P.; writing—original draft preparation, S.J. and R.P.; writing—review and editing, S.P., S.J., M.C., V.L., B.N.d.F. and R.P.; visualization, S.J.; supervision, R.P.; project administration, R.P.; funding acquisition, R.P. All authors have read and agreed to the published version of the manuscript.

**Funding:** This research was funded by the Independent Research Fund Denmark, project "Synthetic Dental Radiography using Generative Artificial Intelligence", grant ID 10.46540/3165-00237B.

**Data Availability Statement:** The original data used for training the models in this study are openly available on Hugging Face at https://huggingface.co/datasets/ibrahimhamamci/DENTEX, accessed on 29 January 2025 or [25]. A GitHub repository containing the WGAN-GP model, source code and synthetic data samples is available at https://github.com/ViktorLaden/DataProjectGAN, accessed on 29 January 2025.

## Appendix A

*Appendix A.1. Dunn's Test*

Dunn's post hoc test was performed on the expert evaluation scores of the four models, following the significant results of the Kruskal–Wallis test. The *p*-values obtained from Dunn's test are summarized in Table A1:

**Table A1.** Dunn's test results (*p*-values) comparing Models 1–4. Values below 0.05 indicate a statistically significant difference in the expert scores between models.

|         | **Model 1** | **Model 2** | **Model 3** | **Model 4** |
|---------|-------------|-------------|-------------|-------------|
| Model 1 | 1.00000000  | 0.85579467  | 0.00003986  | 0.08296606  |
| Model 2 | 0.85579467  | 1.00000000  | 0.00000001  | 0.00051443  |
| Model 3 | 0.00003986  | 0.00000001  | 1.00000000  | 0.24624982  |
| Model 4 | 0.08296606  | 0.00051443  | 0.24624982  | 1.00000000  |

From the statistical evaluation, significant differences were observed between

- **Model 1 and Model 3** ($p = 0.00003986$);
- **Model 2 and Model 3** ($p = 0.00000001$);
- **Model 2 and Model 4** ($p = 0.00051443$).

No significant differences were found between Model 1 and Model 2, Model 1 and Model 4, or Model 3 and Model 4.

## References

1. Pauwels, R. A brief introduction to concepts and applications of artificial intelligence in dental imaging. *Oral Radiol.* **2021**, *37*, 153–160. [CrossRef] [PubMed]
2. Schwendicke, F.A.; Samek, W.; Krois, J. Artificial intelligence in dentistry: Chances and challenges. *J. Dent. Res.* **2020**, *99*, 769–774. [CrossRef]
3. Mörch, C.; Atsu, S.; Cai, W.; Li, X.; Madathil, S.; Liu, X.; Mai, V.; Tamimi, F.; Dilhac, M.; Ducret, M. Artificial intelligence and ethics in dentistry: A scoping review. *J. Dent. Res.* **2021**, *100*, 1452–1460. [CrossRef] [PubMed]
4. Topol, E.J. High-performance medicine: The convergence of human and artificial intelligence. *Nat. Med.* **2019**, *25*, 44–56. [CrossRef]
5. Pauwels, R.; Iosifidis, A. Deep Learning in Image Processing: Part 1—Types of Neural Networks, Image Segmentation. In *Artificial Intelligence in Dentistry*; Springer: Cham, Switzerland, 2024; pp. 283–316.
6. Pauwels, R.; Iosifidis, A. Deep Learning in Image Processing: Part 2—Image Enhancement, Reconstruction and Registration. In *Artificial Intelligence in Dentistry*; Springer: Cham, Switzerland, 2024; pp. 317–351.
7. Hosny, A.; Parmar, C.; Quackenbush, J.; Schwartz, L.H.; Aerts, H.J. Artificial intelligence in radiology. *Nat. Rev. Cancer* **2018**, *18*, 500–510. [CrossRef]
8. Goodfellow, I.; Pouget-Abadie, J.; Mirza, M.; Xu, B.; Warde-Farley, D.; Ozair, S.; Courville, A.; Bengio, Y. Generative adversarial nets. *Adv. Neural Inf. Process. Syst.* **2014**, *27*, 2672–2680.
9. Arjovsky, M.; Chintala, S.; Bottou, L. Wasserstein generative adversarial networks. In Proceedings of the International Conference on Machine Learning. PMLR, Sydney, Australia, 6–11 August 2017; pp. 214–223.
10. Hamamci, I.E.; Er, S.; Simsar, E.; Sekuboyina, A.; Gundogar, M.; Stadlinger, B.; Mehl, A.; Menze, B. Diffusion-based hierarchical multi-label object detection to analyze panoramic dental X-rays. In Proceedings of the International Conference on Medical Image Computing and Computer-Assisted Intervention, Vancouver, BC, Canada, 8–12 October 2023; Springer: Cham, Switzerland, 2023; pp. 389–399.
11. Abdi, A.H.; Kasaei, S.; Mehdizadeh, M. Automatic segmentation of mandible in panoramic X-ray. *J. Med. Imaging* **2015**, *2*, 044003. [CrossRef]
12. Silva, G.; Oliveira, L.; Pithon, M. Automatic segmenting teeth in X-ray images: Trends, a novel data set, benchmarking and future perspectives. *Expert Syst. Appl.* **2018**, *107*, 15–31. [CrossRef]
13. Turosz, N.; Checinski, K.; Checinski, M.; Brzozowska, A.; Nowak, Z.; Sikora, M. Applications of artificial intelligence in the analysis of dental panoramic radiographs: an overview of systematic reviews. *Dentomaxillofac. Radiol.* **2023**, *52*, 20230284. [CrossRef] [PubMed]

14. Kang, S.R.; Shin, W.; Yang, S.; Kim, J.E.; Huh, K.H.; Lee, S.S.; Heo, M.S.; Yi, W.J. Structure-preserving quality improvement of cone beam CT images using contrastive learning. *Comput. Biol. Med.* **2023**, *158*, 106803. [CrossRef]

15. Aetesam, H.; Maji, S.K. Perceptually motivated generative model for magnetic resonance image denoising. *J. Digit. Imaging* **2023**, *36*, 725–738. [CrossRef]

16. Moran, M.B.H.; Faria, M.D.B.; Giraldi, G.A.; Bastos, L.F.; Conci, A. Using super-resolution generative adversarial network models and transfer learning to obtain high resolution digital periapical radiographs. *Comput. Biol. Med.* **2021**, *129*, 104139. [CrossRef] [PubMed]

17. Li, Z.; Cai, A.; Wang, L.; Zhang, W.; Tang, C.; Li, L.; Liang, N.; Yan, B. Promising generative adversarial network based sinogram inpainting method for ultra-limited-angle computed tomography imaging. *Sensors* **2019**, *19*, 3941. [CrossRef] [PubMed]

18. Gao, L.; Xie, K.; Wu, X.; Lu, Z.; Li, C.; Sun, J.; Lin, T.; Sui, J.; Ni, X. Generating synthetic CT from low-dose cone-beam CT by using generative adversarial networks for adaptive radiotherapy. *Radiat. Oncol.* **2021**, *16*, 202. [CrossRef]

19. Nie, D.; Trullo, R.; Petitjean, C.; Ruan, S.; Shen, D. Medical image synthesis with context-aware generative adversarial networks. *arXiv* **2016**, arXiv:1612.05362.

20. Fan, J.; Cao, X.; Wang, Q.; Yap, P.T.; Shen, D. Adversarial learning for mono- or multi-modal registration. *Med. Image Anal.* **2019**, *58*, 101545. [CrossRef]

21. Gulrajani, I.; Ahmed, F.; Arjovsky, M.; Dumoulin, V.; Courville, A.C. Improved training of wasserstein gans. *Adv. Neural Inf. Process. Syst.* **2017**, *30*, 5767–5777.

22. Mao, X.; Li, Q.; Xie, H.; Lau, R.Y.; Wang, Z.; Paul Smolley, S. Least squares generative adversarial networks. In Proceedings of the IEEE International Conference on Computer Vision, Venice, Italy, 22–29 October 2017; pp. 2794–2802.

23. Radford, A. Unsupervised representation learning with deep convolutional generative adversarial networks. *arXiv* **2015**, arXiv:1511.06434.

24. Karras, T. Progressive Growing of GANs for Improved Quality, Stability, and Variation. *arXiv* **2017**, arXiv:1710.10196.

25. Hamamci, I.E.; Er, S.; Simsar, E.; Yuksel, A.E.; Gultekin, S.; Ozdemir, S.D.; Yang, K.; Li, H.B.; Pati, S.; Stadlinger, B.; et al. DENTEX: An Abnormal Tooth Detection with Dental Enumeration and Diagnosis Benchmark for Panoramic X-rays. *arXiv* **2023**, arXiv:2305.19112.

26. Tschumperlé, D.; Deriche, R. Vector-valued image regularization with PDEs: A common framework for different applications. *IEEE Trans. Pattern Anal. Mach. Intell.* **2005**, *27*, 506–517. [CrossRef]

27. Schindelin, J.; Arganda-Carreras, I.; Frise, E.; Kaynig, V.; Longair, M.; Pietzsch, T.; Preibisch, S.; Rueden, C.; Saalfeld, S.; Schmid, B.; et al. Fiji: An open-source platform for biological-image analysis. *Nat. Methods* **2012**, *9*, 676–682. [CrossRef] [PubMed]

28. Heusel, M.; Ramsauer, H.; Unterthiner, T.; Nessler, B.; Hochreiter, S. Gans trained by a two time-scale update rule converge to a local nash equilibrium. *Adv. Neural Inf. Process. Syst.* **2017**, *30*, 6626–6637.

29. Van der Maaten, L.; Hinton, G. Visualizing data using t-SNE. *J. Mach. Learn. Res.* **2008**, *9*, 2579–2605.

30. He, K.; Zhang, X.; Ren, S.; Sun, J. Deep residual learning for image recognition. In Proceedings of the IEEE Conference on Computer Vision and Pattern Recognition, Las Vegas, NV, USA, 27–30 June 2016; pp. 770–778.

31. Zhang, H.; Goodfellow, I.; Metaxas, D.; Odena, A. Self-attention generative adversarial networks. In Proceedings of the International Conference on Machine Learning. PMLR, Long Beach, CA, USA, 9–15 June 2019; pp. 7354–7363.

32. Wang, T.; Liu, M.; Zhu, J.; Tao, A.; Kautz, J.; Catanzaro, B. High-resolution image synthesis and semantic manipulation with conditional GANs. CoRR abs/1711.11585 (2017). *arXiv* **2017**, arXiv:1711.11585.

33. Ledig, C.; Theis, L.; Huszar, F.; Caballero, J.; Cunningham, A.; Acosta, A.; Aitken, A.; Tejani, A.; Totz, J.; Wang, Z.; et al. Photo-realistic single image super-resolution using a generative adversarial network. *arXiv* **2016**, arXiv:1609.04802.

34. Ho, J.; Jain, A.; Abbeel, P. Denoising diffusion probabilistic models. *Adv. Neural Inf. Process. Syst.* **2020**, *33*, 6840–6851.

35. Kingma, D.; Salimans, T.; Poole, B.; Ho, J. Variational diffusion models. *Adv. Neural Inf. Process. Syst.* **2021**, *34*, 21696–21707.

36. Johnson, J.; Alahi, A.; Fei-Fei, L. Perceptual losses for real-time style transfer and super-resolution. In Proceedings of the Computer Vision—ECCV 2016: 14th European Conference, Amsterdam, The Netherlands, 11–14 October 2016; Proceedings, Part II 14; Springer: Cham, Switzerland, 2016; pp. 694–711.

37. Zhang, R.; Isola, P.; Efros, A.A.; Shechtman, E.; Wang, O. The unreasonable effectiveness of deep features as a perceptual metric. In Proceedings of the IEEE Conference on Computer Vision and Pattern Recognition, Salt Lake City, UT, USA, 18–22 June 2018; pp. 586–595.

38. Huang, X.; Belongie, S. Arbitrary style transfer in real-time with adaptive instance normalization. In Proceedings of the IEEE International Conference on Computer Vision, Venice, Italy, 22–29 October 2017; pp. 1501–1510.